\documentclass[runningheads]{llncs}

\usepackage{accv}

\usepackage{accvabbrv}

\usepackage{graphicx}
\usepackage{booktabs}
\usepackage{multirow}
\usepackage{color, colortbl}
\usepackage{caption}
\usepackage{subcaption}
\usepackage{wrapfig}
\usepackage[ruled]{algorithm2e}
\usepackage{algpseudocode}%
\usepackage{amsmath,mathtools}
\usepackage{siunitx}
\usepackage{booktabs, makecell}
\usepackage{floatflt}

\usepackage[accsupp]{axessibility}  

\SetKwComment{Comment}{/* }{ */}

\usepackage[pagebackref,breaklinks,colorlinks,citecolor=accvblue]{hyperref}

\usepackage{orcidlink}
\graphicspath{{figs/}}

\begin{document}

\title{EQ-CBM: A Probabilistic Concept Bottleneck with Energy-based Models and Quantized Vectors}

\titlerunning{EQ-CBM}

\author{Sangwon Kim\inst{1}\orcidlink{0000-0002-7452-3897} \and
  Dasom Ahn\inst{2}\orcidlink{0009-0009-4123-072X} \and
  Byoung Chul Ko\inst{2}\orcidlink{0000-0002-7284-0768} \and
  In-su Jang\inst{1}\orcidlink{0000-0002-0468-4193} \and
  Kwang-Ju Kim \inst{1,}\thanks{Corresponding author.}\orcidlink{0000-0001-8458-4506}}

\authorrunning{S. Kim et al.}

\institute{ETRI, South Korea\\
  \email{\{eddiekim,jef1015,kwangju\}@etri.re.kr}\\
  \and
  Keimyung University, South Korea\\
  \email{tommydasomahn@gmail.com, niceko@kmu.ac.kr}}

\maketitle

\begin{abstract}
  The demand for reliable AI systems has intensified the need for interpretable deep neural networks. Concept bottleneck models (CBMs) have gained attention as an effective approach by leveraging human-understandable concepts to enhance interpretability. However, existing CBMs face challenges due to deterministic concept encoding and reliance on inconsistent concepts, leading to inaccuracies. We propose EQ-CBM, a novel framework that enhances CBMs through probabilistic concept encoding using energy-based models (EBMs) with quantized concept activation vectors (qCAVs). EQ-CBM effectively captures uncertainties, thereby improving prediction reliability and accuracy. By employing qCAVs, our method selects homogeneous vectors during concept encoding, enabling more decisive task performance and facilitating higher levels of human intervention. Empirical results using benchmark datasets demonstrate that our approach outperforms the state-of-the-art in both concept and task accuracy.
\end{abstract}

\section{Introduction}\label{sec:intro}

With the increasing demand for reliable AI, explaining deep neural networks (DNNs) has gained significant attention across various research fields. For decades, post-hoc explanation methods \cite{gradcam,gradcam++,SHAP,LIME,DeepLIFT} have been the mainstay due to their clarity and compatibility. However, these methods fall short in fully interpreting black-box DNNs as they provide explanations detached from the intrinsic decision-making processes \cite{nauta2021neural,ViT-NeT} of models. Recently, concept-based model interpretation has emerged as a promising alternative. Kim \etal \cite{TCAV} define concepts as discriminative vectors necessary for a model to understand objects, presented as orthogonal vectors known as concept activation vectors (CAVs). Building on this, concept bottleneck models (CBMs) \cite{CBM, CEM, ECBM, prob, yuksekgonul2022post, chauhan2023interactive, antehoc, havasi2022addressing, COOP, CGEM} have been proposed to explain the decision-making processes of DNNs using human-understandable concepts. For example, these concepts include \verb|feather color|, \verb|body shape|, or \verb|beak length|, representing distinct attributes of a bird (object) in an image. Various CBMs express these concepts as scores, representing the probability of each concept’s presence in the input. CBMs leverage these interpretable concepts for final task prediction without relying on auxiliary features, thereby enhancing interpretability and transparency.
 
Despite their potential, previous CBMs face limitations in real-world scenarios due to their deterministic concept encoding approach, which uses parametric mappings from input features to concepts. This can lead to inaccuracies, often mapping different concepts to the same latent variables. For instance, CEM \cite{CEM} shows minimal performance changes in human intervention tests without random intervention learning. Moreover, the decision-making process in CBMs can rely on inconsistent concepts, leading to less discriminative concept utilization and impeding effective human intervention. Incorrect concept encoding from complex inputs further constrains task performance.

To address these challenges, we propose a framework that enhances CBMs through probabilistic concept encoding using \textbf{E}nergy-based models (EBMs) with \textbf{Q}uantized concept activation vectors (qCAVs), referred to as \textbf{EQ-CBM}. EBMs enable robust probabilistic inference by modeling the joint energy-concept landscape for each concept. By incorporating qCAVs, our method selects homogeneous vectors during concept encoding, thereby improving task accuracy and interpretability across various images. This enhancement also facilitates greater human intervention, increasing the model’s reliability in complex decision-making processes.

\begin{figure}[h]
  \centering
  \begin{subfigure}[b]{0.4\textwidth}
    \centering
    \includegraphics[height=2.4cm]{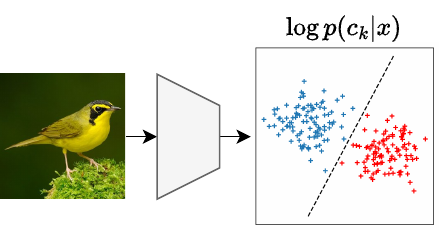}
    \caption{Deterministic Concept Encoding}
    \label{fig:1-1}
  \end{subfigure}
  \hfill
  \begin{subfigure}[b]{0.58\textwidth}
    \centering
    \includegraphics[height=2.4cm]{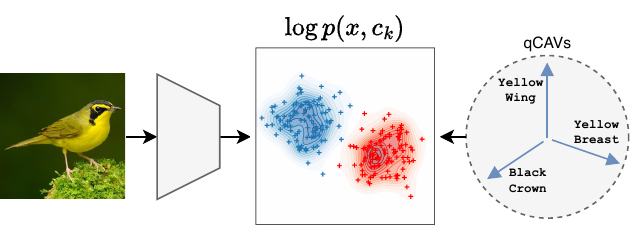}
    \caption{Probabilistic Concept Encoding using qCAVs}
    \label{fig:1-2}
  \end{subfigure}

  \caption{Comparison of concept encoding methods: (a) Deterministic concept encoding, and (b) probabilistic concept encoding using qCAVs.}
  \label{fig:1}

\end{figure}

As illustrated in Fig. \ref{fig:1}, our approach contrasts with previous deterministic concept encoding. In deterministic encoding, concepts are directly mapped from latent vectors to a fixed representation, often leading to less accurate and interpretable results. In contrast, our approach employs energy-based models to infer the relationship between latent vectors and qCAVs, allowing for a more nuanced and probabilistically robust representation of concepts. This approach not only enhances interpretability but also ensures more consistent and reliable concept encoding across various scenarios. Our contributions include:


  \begin{itemize}
    \item Proposing EQ-CBM, a novel framework that enhances CBMs through probabilistic concept encoding using EBMs with qCAVs, addressing the limitations of deterministic encoding.
    \item Introducing a robust decision-making that leverages qCAVs to select relevant concepts, improving task performance and enhancing model interpretability.
    \item Demonstrating the effectiveness of EQ-CBM in both interpretability and accuracy through extensive experiments on multiple datasets, showcasing its superiority over existing CBM approaches.
    \end{itemize}


\section{Backgrounds}\label{sec:2}

\subsection{Concept Bottleneck Models}

CBMs \cite{CBM, CEM, ECBM, prob, yuksekgonul2022post, chauhan2023interactive, antehoc, havasi2022addressing, COOP, CGEM} were designed to enhance the interpretability of DNNs by leveraging human-understandable concepts. These models transform raw input images into a set of intermediate concept representations, which are then used to make final tasks without relying on additional image features. To explain CBMs in detail, we first define the notations. A dataset $\mathcal{D}$ consists of $N$ triplets, each containing an image $x$, a ground truth concept label set $\mathbf{C}^*$, and a ground truth class label $y^*$: $\mathcal{D} = {(x, \mathbf{C}^*, y^*)}^N$. The ground truth concept label set $\mathbf{C}^*$ includes concept labels for $K$ individual concepts, $\mathbf{C}^* \in \{0, 1\}^K$.

CBMs typically comprises three main components:

\begin{enumerate}
\item \textbf{Backbone network} $f: x \rightarrow z$: This network extracts a latent vector $z$ from the input image $x$.
\item \textbf{Concept encoder} $g: z \rightarrow \mathbf{C}$: This encoder maps the latent vector $z$ to a set of concepts $\mathbf{C}$.
\item \textbf{Downstream layer} $h: \mathbf{C} \rightarrow y$: This layer predicts the final class $y$ using only the predicted concepts $\mathbf{C}$.
\end{enumerate}

To maintain structural transparency in our model, we employ a lightweight backbone network, ResNet34 \cite{resnet}, for $f$, and a single fully-connected layer for $h$. This configuration ensures that the model’s decision-making process remains interpretable and transparent. By using these interpretable concepts for final classification tasks, CBMs inherently promote transparency and clarity in model predictions. This transparency allows for human intervention to correct model failures or misalignments between concepts and the final task.

\subsection{Energy-based Models}

EBMs \cite{nijkamp2020anatomy, gao2018learning, zhao2016energy, du2021unsupervised, pang2020learning, du2019implicit, han2020joint, grathwohl2019your, kim2022energy, yang2021jem++, guo2023egc} are probabilistic frameworks designed to represent complex distributions. The core principle of EBMs is to associate an energy score with each possible state of variables, where lower energy scores correspond to more probable states. This approach provides a flexible and expressive means to capture the underlying dependencies within the data.

EBMs employ an energy function $E_\theta(x)$, which is parameterized by a compact multi-layer perceptron and maps each variable state to a scalar energy score. This function encapsulates the interactions and dependencies among the variables. More likely states are assigned lower energy scores, while less likely states receive higher scores. The probability of $x$ in EBMs is determined by Boltzmann distribution:

\begin{equation}
  p_\theta(x) = \frac{\mathrm{exp}(-E_\theta(x))}{Z(\theta)}, \quad Z(\theta)=\int_{x} \mathrm{exp}(-E_\theta(x)),
\end{equation}

\noindent
where $Z(\theta)$ is a partition function, which is crucial for normalizing the distribution. It sums the contributions of all possible states, allowing the distribution to be properly normalized. However, calculating $Z(\theta)$ directly is often intractable due to the vast number of possible states.

To address this, Kullback-Leibler (KL) divergence is typically minimized to approximate $p_\theta(x)$ to the true data distribution $p_\mathcal{D}(x)$. This involves maximizing the expected log-likelihood of $p_\theta$:

\begin{equation}
  \underset{\theta}{\mathrm{max}} \mathbb{E}_{p_\mathcal{D}} [\log p_\theta (x)]
\end{equation}

The gradient of the log-likelihood w.r.t. the parameters $\theta$ is given by:

\begin{equation}
  \frac{\partial \log p_\theta (x)}{\partial \theta} = \mathbb{E}_{p_\theta (x')} \bigl[\frac{\partial E_\theta (x')}{\partial \theta}\bigr] - \frac{\partial E_\theta (x)}{\partial \theta}
\end{equation}


This gradient can be derived as follows:

\begin{align}
  \log p_\theta (x) &= \log \frac{\mathrm{exp}(-E_\theta(x))}{Z(\theta)} = \log [\mathrm{exp}(-E_\theta (x))] - \log [Z(\theta)]\\
   &= -\log Z(\theta) - E_\theta (x) \label{eq:6}
\end{align}

\begin{align}
   \nabla_\theta \log p_\theta (x) &= - \frac{1}{Z(\theta)}\nabla_\theta Z(\theta) - \nabla_\theta E_\theta (x)\\
   &= - \frac{1}{Z(\theta)} \nabla_\theta \int_x \mathrm{exp}(-E_\theta (x)) - \nabla_\theta E_\theta (x)\\
   &= \frac{1}{Z(\theta)} \int_{x'} \exp (-E_\theta (x')) - \nabla_\theta E_\theta (x)\\
   &= \int_x' \frac{\exp (-E_\theta (x'))}{Z(\theta)} \nabla_\theta E_\theta (x') - \nabla_\theta E_\theta (x)\\
   &= \mathbb{E}_{p_\theta(x')}[\nabla_\theta E_\theta(x')] - \nabla_\theta E_\theta(x) \label{eq:11}
\end{align}

Direct sampling from $p_\theta(x)$ to obtain negative sample $x'$ is often impractical due to the intractability of the partition function  $Z(\theta)$. To address this challenge, sampling methods such as Markov Chain Monte Carlo (MCMC) \cite{nijkamp2020anatomy, nijkamp2019learning, nijkamp2020learning, han2017alternating} are utilized. MCMC facilitates approximate inference by sampling from the distribution without the need to explicitly compute $Z(\theta)$.

Previous works have employed methods such as Gibbs sampling \cite{ackley1985learning, hinton2006fast, salakhutdinov2009deep} and Stochastic Gradient Langevin Dynamics (SGLD) \cite{neal2011mcmc, zhu1998grade} to approximate the true distribution, enabling efficient learning and inference in EBMs. SGLD combines gradient descent with Langevin dynamics to sample from the distribution. The SGLD updates are given by:

\begin{equation}
  x_0 \sim p_0(x),\quad x_{t+1} = x_t - \frac{\gamma}{2} \frac{\partial E_\theta(x_t)}{\partial x_t} + \epsilon, \quad \epsilon \sim \mathcal{N}(0, \gamma)\label{eq:12}
\end{equation}

\noindent
where $\gamma$ is the step size and $\epsilon$ is Gaussian noise. Learning in EBMs involves adjusting the parameters of the energy function to minimize the energy of states corresponding to the observed data. This can be expressed as minimizing the log-likelihood (Eq. \ref{eq:6}). Since $\log Z(\theta)$ is intractable, SGLD is used to approximate the gradient of the log-likelihood w.r.t. the model parameters (Eq. \ref{eq:11}).

In this paper, we leverage EBMs within the concept encoder to probabilistically infer the relationship between qCAVs and the variational latent vector $v_k$. By performing probabilistic relation modeling, we can obtain robust concepts that are more resilient to the variabilities and complexities of real-world scenarios.

\section{EQ-CBM}\label{sec:3}

\begin{figure}[t]
  \centering
  \includegraphics[width=0.95\textwidth]{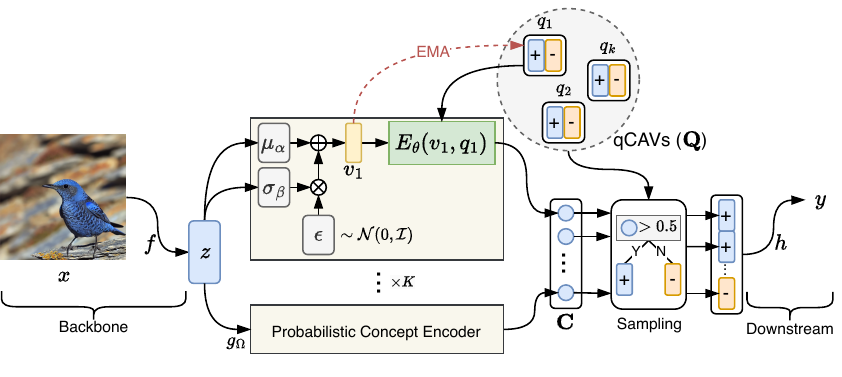}
  \caption{Overall architecture of the EQ-CBM. The input image $x$ is processed by the backbone network $f$ to generate a latent vector $z$. This latent vector is fed into probabilistic concept encoders $g_\mathrm{\Omega}$, which use variational inference techniques to infer $v_k$. The energy function $E_\theta$ evaluates the compatibility of $v_k$ with the qCAVs. Exponential Moving Average (EMA) updates each vector pair. The sampling module selects the concept vectors having lowest energy scores, which are then used for the final task.}
  \label{fig:2}
\end{figure}

We introduce EQ-CBM, a framework that enhances CBMs by utilizing probabilistic concept encoding with EBMs with qCAVs. Our approach addresses the limitations of previous CBMs by improving robustness and interpretability in complex decision-making tasks. As depicted in Fig. \ref{fig:2}, the proposed method comprises three main components: qCAVs ($\mathbf{Q}$), probabilistic concept encoders ($g_\mathrm{\Omega}$), and a sampling module.

In this section, we describe each component and its integration into the overall framework, highlighting the advantages of using energy-based modeling and vector quantization for concept representation.



\subsection{Quantized Concept Activation Vectors}

Vector quantization is a powerful technique widely used to discretize continuous vectors into a finite set of representative vectors, known as a codebook. Notable applications of vector quantization include VQ-VAE \cite{vqvae}, which avoids posterior collapse and enables high-quality generative modeling. By mapping continuous data to discrete codebook vectors, vector quantization captures the underlying structure of the data by aligning representations with a finite set of learned codebook vectors. This alignment is particularly beneficial for interpretability and robustness, as it simplifies the analysis and manipulation of the model’s internal representations.

In our approach, we propose qCAVs to enhance the interpretability and robustness of CBMs. This discretization helps in standardizing the learned concepts, facilitating more straightforward model behavior and enabling effective human intervention. Moreover, quantized concepts provide a stable and consistent basis for decision-making, improving the model’s reliability in complex real-world scenarios.

As shown in Fig. \ref{fig:2}, qCAVs consist of $K$ non-differentiable vector pairs: $\mathbf{Q}=\{(q^+_k, q^-_k)=q_k\}_{\in [K]}$, $q_k^+ \in \mathbb{R}^d$ and $q_k^- \in \mathbb{R}^d$. Each pair represents a positive and negative vector for a particular $k^{\mathrm{th}}$ concept. In our probabilistic concept encoder, these vectors are used as conditioning variables to define a joint energy-concept landscape. The detailed process for updating qCAVs is described in the following section.


\subsection{Probabilistic Concept Encoder}


The probabilistic concept encoders $g_\mathrm{\Omega}$ abstract each concept by applying variational inference techniques to extract a variational latent vector $v_k$ from a normal distribution. Specifically, as depicted in Fig. \ref{fig:2}, from $z$, we infer the mean $\mu_\alpha$ and variance $\sigma_\beta$, and sample noise $\epsilon$ from Gaussian distribution to learn diverse representations. This process enables the model to capture the variability and complexity of real-world concepts more effectively.

Our energy function $E_\theta$ then models the joint distribution between the variational latent vector $v_k$ and the qCAVs. Instead of updating qCAVs through backpropagation gradients, we use an exponential moving average (EMA) to update each vector pair, which is more effective for ensuring stability and consistent updates. The EMA is applied as follows:

\begin{align}
  q_{k,s}^{+} :=\begin{cases}
    q_{k, s-1}^{+} \cdot \eta + v_k \cdot (1-\eta)  & c^*_k = 1\\
    q_{k, s-1}^{+} & otherwise\\  
  \end{cases}\label{eq:emap}
\end{align}

\begin{align}
  q_{k, s}^{-} :=\begin{cases}
    q_{k, s-1}^{-} \cdot \eta + v_k \cdot (1-\eta)  & c^*_k = 0\\
    q_{k, s-1}^{-} & otherwise\\  
  \end{cases}\label{eq:eman}
\end{align}

\noindent
where $s$ represents the training steps and $\eta$ is the decay factor, set to 0.95, ensuring a balance between the historical values of the qCAVs and the newly observed variational latent vectors $v_k$. This balance allows the model to adapt gradually without abrupt changes.

\subsection{Energy-based Concept Encoding}

\begin{wrapfigure}{r}{0.5\textwidth}
  \centering
    \includegraphics[width=0.49\textwidth]{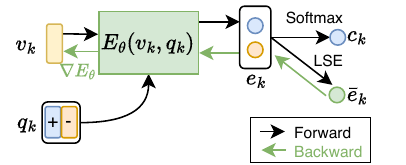}
  \caption{Integration of EBM in the concept encoder. The output is processed through softmax for concept prediction $c_k$ and LogSumExp (LSE) for composed energy score $\bar{e}_k$. The backward pass using SGLD updates $v_k$ based on the joint energy-concept landscape.}%
  \label{fig:3}%
\end{wrapfigure}

To effectively capture the relationships between variational latent vectors $v_k$ and qCAVs, we employ an energy-based approach for concept encoding. EBMs offer a robust and flexible framework for modeling complex dependencies and probabilistic relationships. By utilizing an energy function $E_\theta$, we measure the compatibility between $v_k$ and predefined qCAVs, enabling a more comprehensible and interpretable representation of concepts.

The core idea behind energy-based concept encoding is to define a joint energy-concept landscape where low energy scores indicate high compatibility between $v_k$ and the qCAVs. This approach allows the model to probabilistically infer concept activations given $v_k$, thereby effectively capturing the variability and complexity of real-world scenarios. As shown in Fig. \ref{fig:3}, we integrate EBMs into the concept encoders, which operates on the variational latent vector $v_k$. This energy function returns a low value when $v_k$ is closely related to the $q_k$, facilitating the selection of the most relevant concept vectors for the final task. The problem we aim to solve is therefore represented as the joint distribution of $c_k$, $v_k$, and $q_k$ as follows:


\begin{equation}
  \log p_\mathrm{\Omega} (c_k, v_k, q_k) = \log p_\theta(v_k, q_k) + \log p_\mathrm{\Omega} (c_k | v_k, q_k),
\end{equation}

\noindent
where $p_\mathrm{\Omega} (c_k | v_k, q_k)$ is normalized w.r.t. $c_k$, making it straightforward to compute by maximizing the likelihood. However, since EBMs are unnormalized, directly maximizing the likelihood is more challenging. Therefore, we estimate the gradient of the following likelihood:

\begin{equation}
  \begin{split}
    \nabla_\mathrm{\Omega} \mathbb{E}_{p_\mathcal{D} (c_k, v_k, q_k)} &[\log p_\mathrm{\Omega} (c_k, v_k, q_k)] = \nabla_\mathrm{\Omega} \mathbb{E}_{p_\mathcal{D}(c_k, v_k, q_k)}[\log p_\mathrm{\Omega} (c_k|v_k, q_k)] \\&+ \nabla_\theta \mathbb{E}_{p_\theta (v_k', q_k)}[\nabla_\theta E_\theta (v_k', q_k)] - \nabla_\theta \mathbb{E}_{p_\mathcal{D}(v_k, q_k)}[\nabla_\theta E_\theta (v_k, q_k)]  ,
  \end{split}\label{eq:15}
\end{equation}

\noindent
where the first term is equivalent to cross-entropy and can be treated similarly to a standard classification task conditioned on $q_k$. The second term, $\nabla_\theta \mathbb{E}_{p_\theta (v_k', q_k)}[\nabla_\theta E_\theta (v_k', q_k)]$, forms gradients that increase the energy function $E_\theta (v_k', q_k)$ for the negative samples $v_k'$. Conversely, the third term reduces the energy for the actual samples $v_k$, learning to minimize the energy for these real samples.

Following previous EBMs \cite{EBM, han2020joint, kim2022energy}, we use SGLD to synthesize negative samples for the second term in Eq. \ref{eq:12}. SGLD generates samples that approximate the true distribution. To facilitate this, we redefine the problem as a binary classification task to determine whether the variational latent vector $v_k$ is related to either $q_k^+$ or $q_k^-$. Consequently, the energy function is defined as $E_\theta:(v_k, q_k) \rightarrow e_k \in \mathbb{R}^2$. Unlike prior approaches that utilized a single scalar value for energy scores, we interpret the energy function as class logits. By employing the LogSumExp (LSE) function, as shown in Eq. \ref{eq:17}, the model concurrently addresses both classification and energy modeling, inspired by the work of Grathwohl \etal \cite{grathwohl2019your}. The update rule for SGLD is as follows. For simplicity, we omit the indexing $k$:

\begin{algorithm}[t]
  \scriptsize

      \DontPrintSemicolon
      \caption{Learning algorithm}\label{alg:1}
    \KwIn{image $x$, ground truth concepts $\mathbf{C}^*=\{c^*_1, \dots, c^*_k\}$, qCAVs $\mathbf{Q}=\{q_1, \dots, q_k\}$}

    $\mathbf{C} \gets \emptyset$\;
    $\bar{e}, \,\bar{e}' \gets 0$\;
    $z \gets f(x)$ \Comment*[r]{Backbone network}

    \ForEach{$\{\alpha, \beta, \theta\}^K \subseteq \mathrm{\Omega}$}{
      $v_k \gets \mu_\alpha(z)+\sigma_\beta(z)\cdot\epsilon$ \Comment*[r]{Variational inference}
      $e_k \gets E_\theta(v_k, q_k)$\;  
      $\bar{e} \gets \bar{e} + \mathrm{LSE}(e_k)$\;

      $v'_k \gets \mathcal{N}(0, \mathcal{I})$\;
      
      \For{$t=1 \ \mathbf{to}\  T$}{
        $v'_k \gets \mathrm{SGLD}(v'_k, q_k, c^*_k)$ \Comment*[r]{Eq. \ref{eq:17}}
      }
      $e'_k \gets E_\theta(v'_k, q_k)$\;
      $\bar{e}' \gets \bar{e}' + \mathrm{LSE}(e'_k)$\;

      $\mathbf{C} \gets \mathbf{C} \cup \mathrm{Softmax}(e_k)[+]$ \Comment*[r]{Obtain the concept score for $q_k^+$}
      $q_k \gets \mathrm{EMA}( v_k, q_k, c^*_k)$ \Comment*[r]{Eqs. \ref{eq:emap}-\ref{eq:eman}}

    }

    $y \gets h(\mathrm{sampling}(\mathbf{Q},\mathbf{C}))$\;


    \end{algorithm}

\begin{equation}
  v_{0} \sim p_0(v),\quad v_{t+1} = v_{t} - \frac{\gamma}{2} \frac{\partial \log \sum_{c} \exp(E_\theta (v_{t}, q)[c])}{\partial v_{t}} + \epsilon, \quad \epsilon \sim \mathcal{N}(0, \gamma)\label{eq:17}
\end{equation}

To train the energy function, we apply contrastive divergence loss, as represented in Eq. \ref{eq:15}, which offers several benefits. Firstly, it enables the model to learn robust representations by contrasting the energy of positive samples against negative samples. Secondly, it regularizes the model by preventing the energy values from diverging. The energy-based objective function is defined as follows:

\begin{equation}
  \mathcal{L}_e = \frac{1}{K}\sum_k \left(\bar{e}'_k - \bar{e}_k + (\bar{e}'_k + \bar{e}_k)^2 \right)
\end{equation}

\begin{equation}
  \bar{e}'_k = -\mathrm{LSE}\left(E_\theta(v_k', q_k)\right), \quad \bar{e}_k = -\mathrm{LSE}\left(E_\theta(v_k, q_k)\right),
\end{equation}


The overall learning process for the proposed method is summarized in Algorithm \ref{alg:1}, with the final objective function for training the proposed model defined as follows:

\begin{equation}
  \mathcal{L}_{total} = \lambda_c\mathcal{L}_{\mathrm{CE}}(\mathbf{C}, \mathbf{C}^*) + \lambda_y\mathcal{L}_{\mathrm{CE}}(y, y^*) + \lambda_e\mathcal{L}_e
\end{equation}

\section{Experiments}\label{sec:4}

\begin{wraptable}{r}{0.4\textwidth}

  \centering
\scriptsize
\caption{Hyperparameters.}\label{tab:hy}
        \begin{tabular}[b]{lccccccc}
          \toprule
            Datasets & $d$ & $T$ & $\gamma$ & $\lambda_c$ & $\lambda_y$ & $\lambda_e$ & LR\\
            \midrule
            CUB & 16 & 20 & 0.4 &5 &1 &0.05 &0.005\\
            CelebA& 16 & 20 &0.4 & 5&1&0.05 &0.01\\
            AwA2& 16 & 20 & 0.4& 5&1 & 0.05 &0.005\\
            \bottomrule
        \end{tabular}
        
      \end{wraptable}

To evaluate our model, we used four datasets: CUB \cite{welinder2010caltech}, CelebA \cite{CelebA}, AwA2 \cite{xian2018zero}, and TravelingBirds \cite{CBM}. The CUB dataset includes 12K bird images across 200 species, with 6K for training and 6K for testing, each labeled with 312 attributes. Following previous works \cite{CBM, CEM, prob}, we used 112 attributes as concepts. CelebA contains 202K facial images of 10K celebrities, each with 40 attributes, from which we used six attributes as concepts. The AwA2 dataset has 37K images of 50 animal species, annotated with 85 attributes. The TravelingBirds dataset, derived from the CUB dataset \cite{welinder2010caltech}, replaces image backgrounds with diverse scenes \cite{Places} to test model robustness in real-world uncertainties. All images were resized to 299$\times$299 pixels across datasets. All experiments were conducted on a system with an AMD 5955WX CPU and an Nvidia A6000 GPU using five random seeds. Detailed hyperparameters are depicted in Table \ref{tab:hy}.

\subsection{Metrics}

To comprehensively evaluate the performance, we utilized several key metrics: Concept Accuracy, Task Accuracy, and Uncertainty. Each metric provides insight into different aspects of model performance, from the precision of concept predictions to the robustness of the model under uncertain conditions.

\noindent
\textbf{Concept Accuracy} measures how accurately the model predicts predefined concepts. High concept accuracy indicates effective learning and prediction of concepts.

\noindent
\textbf{Task Accuracy} measures the correctness of the model's primary classification task. High task accuracy means the model performs well in predicting the correct class using only concept.

\noindent
\textbf{Uncertainty} assesses the model's confidence and robustness to variability in the data. It is calculated as follows:


\begin{equation}
  e_k = (e_k^+, e_k^-) = E_\theta(v_k, q_k), \quad u_k = \bigl({\exp\bigl(\mathrm{LSE}(e_k) - \mathrm{Mean}(e_k)\bigr) - 1}\bigr)^{-1},
\end{equation}

\noindent
where $(e_k^+, e_k^-)$ are the paired energy scores for a concept $k$, and $u_k$ represents the uncertainty, with higher values indicating greater uncertainty.


\subsection{Concept and Task accuracy}

To evaluate the effectiveness of our proposed model, we compared its performance against several CBMs using three datasets: CUB, CelebA, and AwA2. We measured both concept prediction accuracy and task (classification) accuracy. Table \ref{tab:1} presents the results of these experiments, showing the mean accuracy and confidence intervals (CI) for each method.

\begin{table}[t]
  \centering
  \caption{Comparison of concept and task accuracy across various CBM models (\textbf{Bold}: best score, \underline{underline}: second-best score).}
  \label{tab:1}
  \resizebox{0.98\textwidth}{!}{
    \begin{tabular}{lllllll}
      \toprule
      \multirow{2}{*}{Methods}        & \multicolumn{2}{l}{CUB \cite{CUB}}           & \multicolumn{2}{l}{CelebA \cite{CelebA}}     & \multicolumn{2}{l}{AwA2 \cite{xian2018zero}}                                                                                                                                                                  \\
                                      & Concept \scriptsize{($\pm$CI)}               & Task \scriptsize{($\pm$CI)}                  & Concept \scriptsize{($\pm$CI)}               & Task \scriptsize{($\pm$CI)}                  & Concept \scriptsize{($\pm$CI)}               & Task \scriptsize{($\pm$CI)}                  \\
      \midrule
      Fuzzy-CBM \cite{CBM}            & 95.882 \scriptsize{($\pm$0.105)}             & 74.228 \scriptsize{($\pm$0.606)}             & 90.269 \scriptsize{($\pm$0.211)}             & 33.696 \scriptsize{($\pm$2.104)}             & 99.000 \scriptsize{($\pm$0.168)}             & 95.089 \scriptsize{($\pm$1.005)}             \\
      Bool-CBM \cite{CBM}             & 96.229 \scriptsize{($\pm$0.031)}             & 72.512 \scriptsize{($\pm$0.466)}             & 90.329 \scriptsize{($\pm$0.164)}             & 33.915 \scriptsize{($\pm$0.885)}             & 99.001 \scriptsize{($\pm$0.188)}             & 94.869 \scriptsize{($\pm$1.047)}             \\
      CEM  \cite{CEM}                 & 96.160 \scriptsize{($\pm$0.157)}             & 79.029 \scriptsize{($\pm$0.519)}             & 90.237 \scriptsize{($\pm$0.306)}             & \underline{42.618} \scriptsize{($\pm$1.412)} & \underline{99.048} \scriptsize{($\pm$0.037)} & 95.745 \scriptsize{($\pm$0.294)}             \\
      Prob-CBM  \cite{prob}           & 95.596 \scriptsize{($\pm$0.061)}             & 76.265 \scriptsize{($\pm$0.145)}             & 89.272 \scriptsize{($\pm$0.238)}             & 34.472 \scriptsize{($\pm$0.893)}             & 98.283 \scriptsize{($\pm$0.065)}             & 92.485 \scriptsize{($\pm$0.315)}             \\
      Coop-CBM  \cite{COOP}           & 89.892 \scriptsize{($\pm$0.649)}             & \underline{79.154} \scriptsize{($\pm$0.734)} & \underline{90.534} \scriptsize{($\pm$0.142)} & 42.393 \scriptsize{($\pm$1.354)}             & 98.875 \scriptsize{($\pm$0.107)}             & \underline{95.927} \scriptsize{($\pm$0.153)} \\
      ECBM  \cite{ECBM}               & \underline{96.536} \scriptsize{($\pm$0.091)} & 77.148 \scriptsize{($\pm$0.695)}             & 90.006 \scriptsize{($\pm$0.986)}             & 34.976 \scriptsize{($\pm$2.111)}             & 98.909 \scriptsize{($\pm$0.037)}             & 94.555 \scriptsize{($\pm$0.121)}             \\
      \rowcolor{lightgray}EQ-CBM (Ours) & \bfseries96.580 \scriptsize{($\pm$0.043)}    & \bfseries79.310 \scriptsize{($\pm$0.272)}    & \bfseries90.617 \scriptsize{($\pm$0.309)}    & \bfseries56.600 \scriptsize{($\pm$1.060)}    & \bfseries99.129 \scriptsize{($\pm$0.022)}    & \bfseries95.965 \scriptsize{($\pm$0.102)}    \\
      \bottomrule
    \end{tabular}
  }

\end{table}

As shown in Table \ref{tab:1}, our proposed EQ-CBM outperforms existing approaches in both concept and task accuracy across all three datasets. For the CUB dataset, EQ-CBM achieved a concept accuracy of 96.580\% and a task accuracy of 79.310\%. In comparison, the best-performing baseline, ECBM, achieved a concept accuracy of 96.536\% but a lower task accuracy of 77.148\%. Similarly, for the CelebA dataset, our model attained concept and task accuracies of 90.617\% and 56.600\%, respectively, significantly surpassing the best-performing baseline, CEM, which had a task accuracy of 42.618\%. In the AwA2 dataset, EQ-CBM reached concept and task accuracies of 99.129\% and 95.965\%, respectively. The closest competitor, Coop-CBM, achieved 98.875\% in concept accuracy and 95.927\% in task accuracy.

These results demonstrate that our model not only maintains high concept accuracy but also achieves superior task performance, validating the effectiveness of incorporating EBMs with qCAVs. Notably, our approach consistently outperformed other models in task accuracy, particularly on the CelebA dataset, where it exceeded the second-best method by a substantial margin (56.600\% vs. 42.618\%). This highlights the robustness and generalization capability of our model across different datasets and tasks. The significant improvements in task accuracy, especially on challenging datasets such as the CelebA dataset, underscore the effectiveness of qCAVs in capturing complex relationships between concepts and improving overall model performance. These findings suggest that EQ-CBM can be a valuable addition to existing concept-based models, providing both enhanced interpretability and higher task accuracy.

\subsection{Concept Intervention}

\begin{figure}[t]
  \centering
  \includegraphics[width=0.95\textwidth]{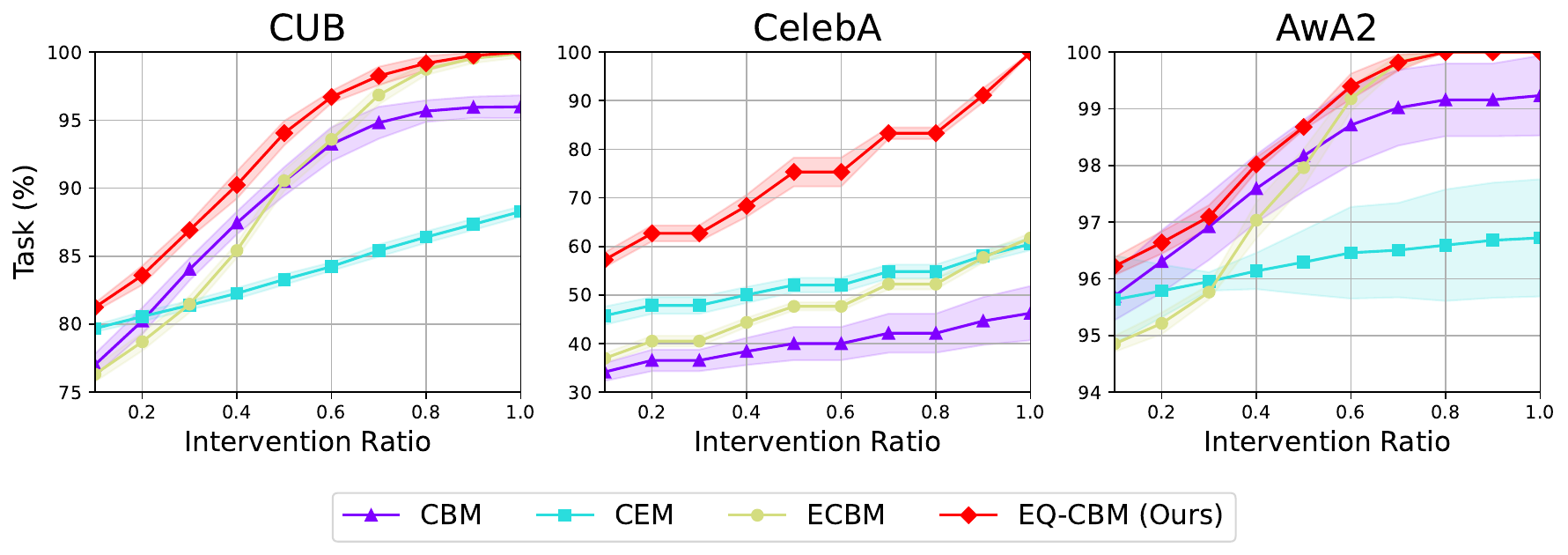}
  \caption{Task accuracy under varying levels of human intervention in concept predictions for different models.}
  \label{fig:4}
\end{figure}

To further evaluate the robustness and interpretability of our model, we conducted a series of concept intervention experiments. The goal of these experiments is to assess how human intervention in correcting concept predictions influences the overall task accuracy. As illustrated in Fig. \ref{fig:4}, we compared the performance of our model (EQ-CBM) with several methods, including CBM, CEM, and ECBM, under varying levels of intervention. The intervention ratio on the x-axis represents the proportion of corrected concept predictions, while the y-axis shows the corresponding task accuracy.

For the CUB dataset, as the intervention ratio increased, EQ-CBM consistently outperformed other models, achieving nearly 100\% task accuracy at higher intervention ratios. This demonstrates the effectiveness of our model in leveraging homogeneous concepts from qCAVs to enhance human intervention. Similarly, in the CelebA dataset, our model showed significant improvements in task accuracy with increasing intervention ratios, outperforming all baseline methods. Notably, EQ-CBM achieved a task accuracy of 100\% with full intervention, highlighting its robustness and capacity to incorporate human feedback effectively. In the AwA2 dataset, EQ-CBM again outperformed the baseline methods across all intervention levels. With an intervention ratio of 1.0, our model reached a task accuracy of 100\%, indicating its superior ability to utilize qCAVs for improved the level of human intervention.

These intervention experiments underscore the advantages of our approach in terms of both robustness and interpretability. By allowing for human intervention in concept predictions, EQ-CBM can significantly enhance the overall accuracy of the model. This is particularly beneficial in real-world applications where model predictions can be iteratively refined through human expertise.

\subsection{Visualization of Encoded Concepts}

\begin{figure}[t]
  \centering
  \begin{subfigure}[b]{0.28\textwidth}
    \centering
    \includegraphics[height=3.3cm]{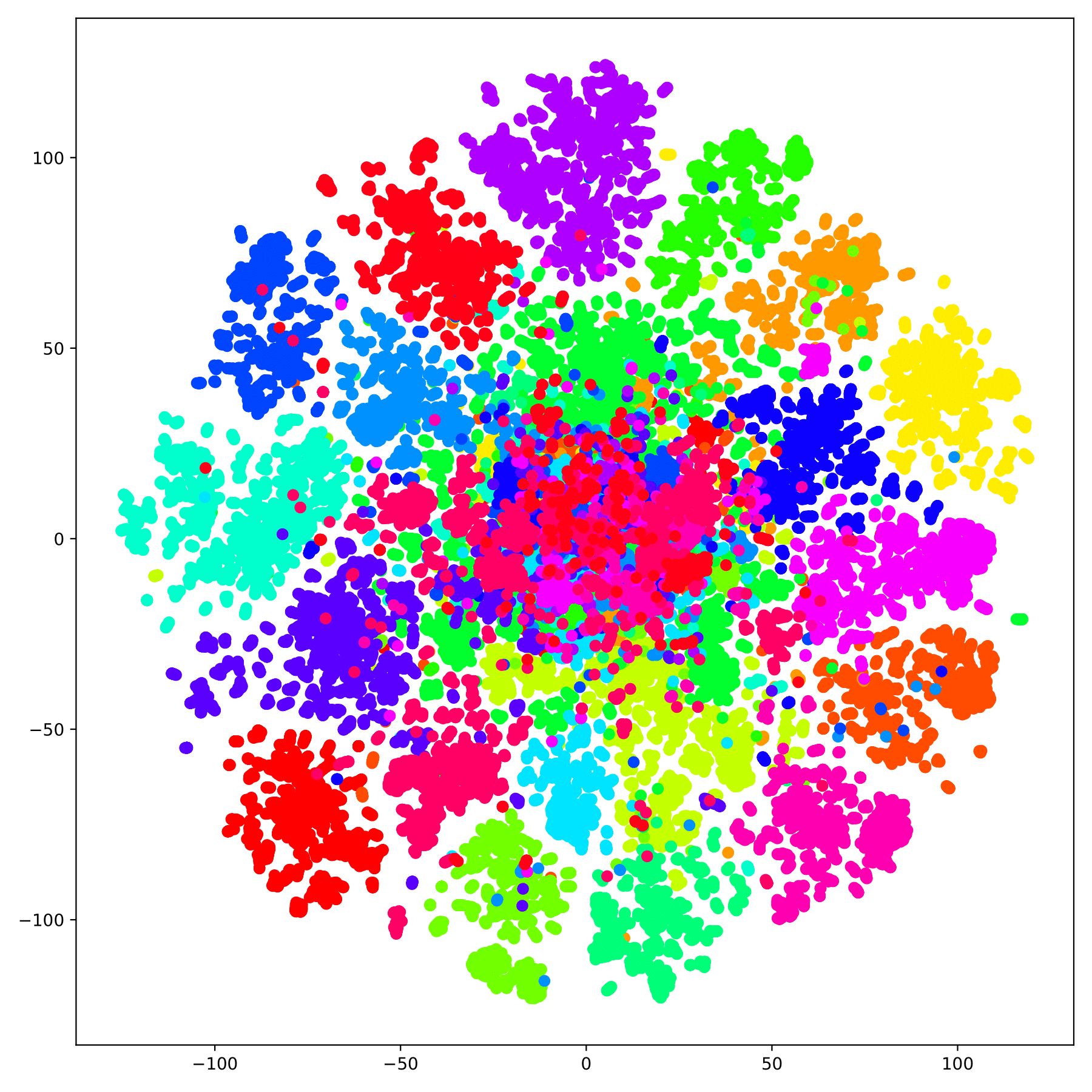}
    \caption{CEM \cite{CEM}}
    \label{fig:5-cem}
  \end{subfigure}
  \hfill
  \begin{subfigure}[b]{0.28\textwidth}
    \centering
    \includegraphics[height=3.3cm]{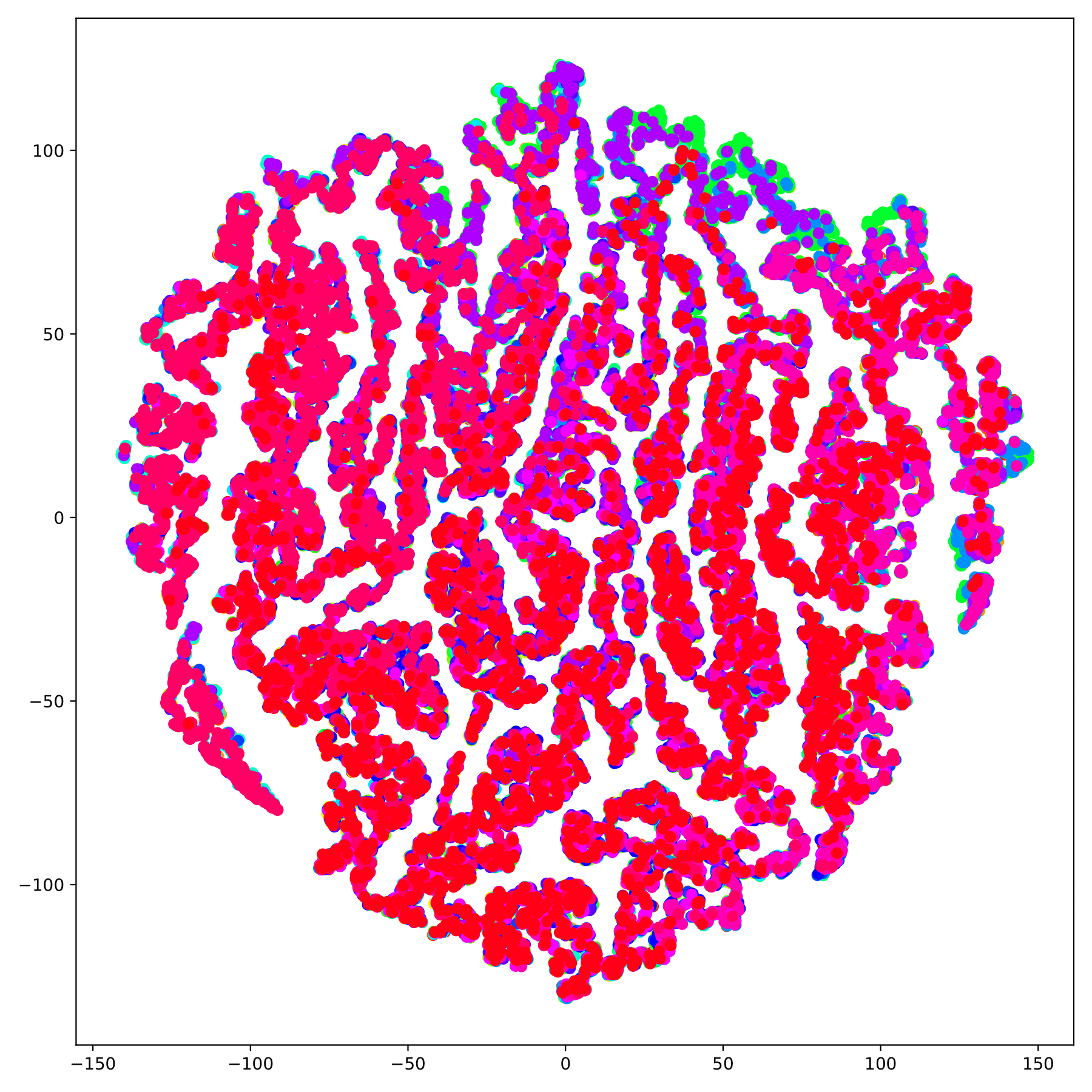}
    \caption{ECBM \cite{ECBM}}
    \label{fig:5-ecbm}
  \end{subfigure}
  \hfill
  \begin{subfigure}[b]{0.42\textwidth}
    \centering
    \includegraphics[height=3.3cm]{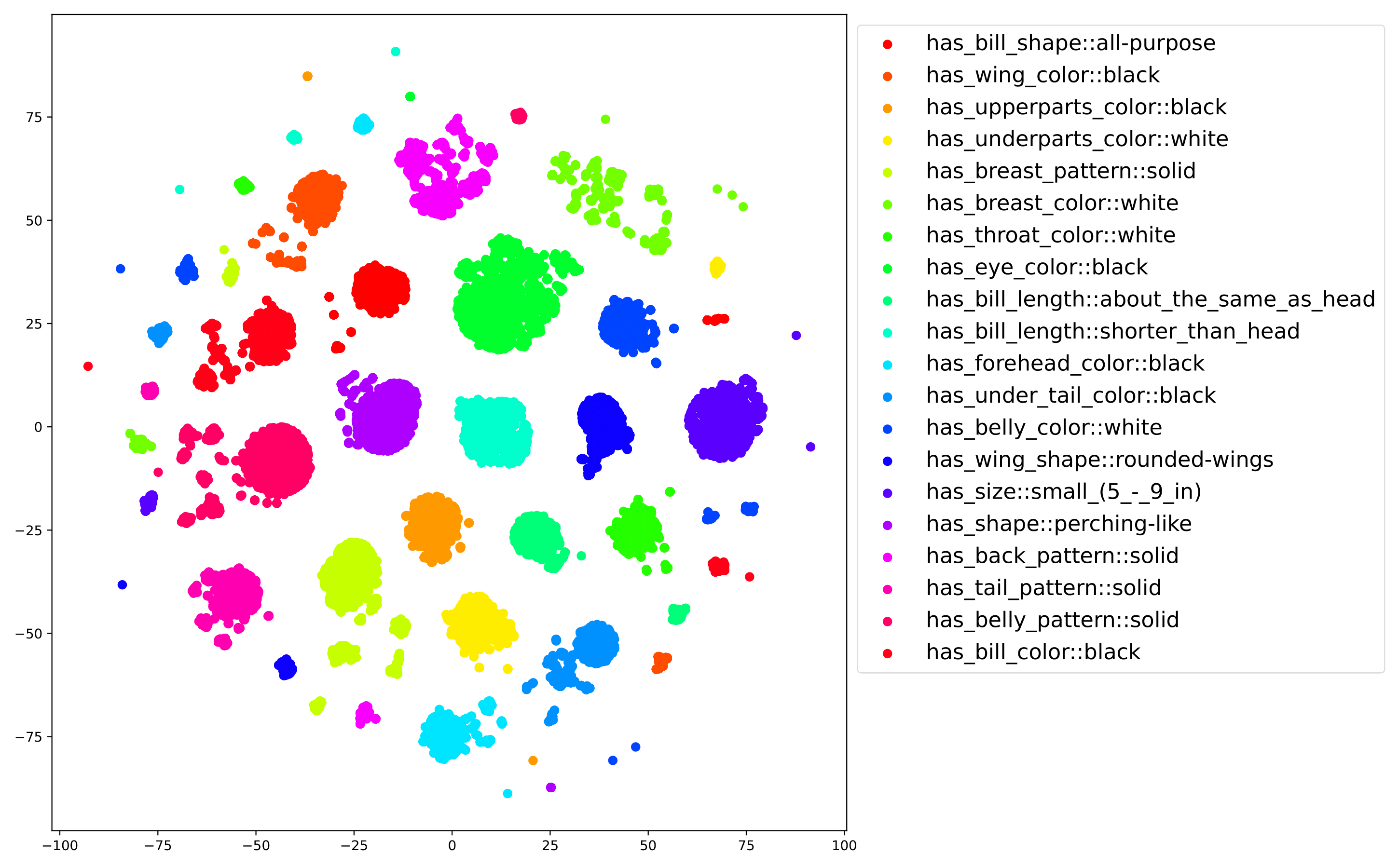}
    \caption{EQ-CBM (Ours)}
    \label{fig:5-ours}
  \end{subfigure}
  \caption{t-SNE visualizations of encoded concepts for (a) CEM, (b) ECBM, and (c) EQ-CBM (Ours). Different colors represent different concepts.}
  \label{fig:5}
\end{figure}

To assess how well the encoded concepts capture the underlying data structure, we utilized t-SNE to visualize the concepts produced by each model just before the final task linear classifier. Figure \ref{fig:5} shows the t-SNE plots of these concepts for CEM \cite{CEM}, ECBM \cite{ECBM}, and our proposed EQ-CBM on the CUB dataset.

In Fig. \ref{fig:5-cem}, the encoded concepts produced by CEM exhibited significant overlap among different concepts, indicating weak separation and potential confusion in concept interpretation. This overlap suggests that the model might struggle to differentiate between certain concepts, leading to less accurate and interpretable predictions. In Fig. \ref{fig:5-ecbm}, while ECBM achieves better separation than CEM, there is still considerable clustering within certain concept groups. This partial overlap could impact the model’s ability to clearly distinguish between similar concepts, potentially affecting its task performance and interpretability. In contrast, our proposed EQ-CBM, as depicted in Fig. \ref{fig:5-ours}, demonstrated a much clearer separation of concepts. The t-SNE plot reveals distinct clusters for each concept, indicating that EQ-CBM effectively captures the unique characteristics of each concept and represents them in a well-separated latent space. Specifically, these concepts are sampled qCAVs, which provide a homogeneous representation of concepts by incorporating variability and uncertainty. This clear separation enhances the model’s interpretability and allows for more accurate concept-based predictions.

\subsection{Uncertainty Robustness}



To evaluate the robustness of our proposed model under uncertain conditions, we conducted experiments using the TravelingBirds dataset, a variant of the CUB dataset where bird image backgrounds are replaced with various real-world scenes from the Places dataset \cite{Places}. This introduced background variability, challenging the model’s ability to maintain performance under these uncertain conditions. The models were trained on the CUB dataset and tested on two versions of the TravelingBirds dataset: CUB\_Black, where backgrounds are replaced with black (Fig. \ref{fig:tbs-a}), and CUB\_Random, where backgrounds are replaced with random real-world scenes (Fig. \ref{fig:tbs-b}).

\begin{figure}
  \centering
  \begin{subfigure}{0.4\linewidth}
    \includegraphics[width=\linewidth]{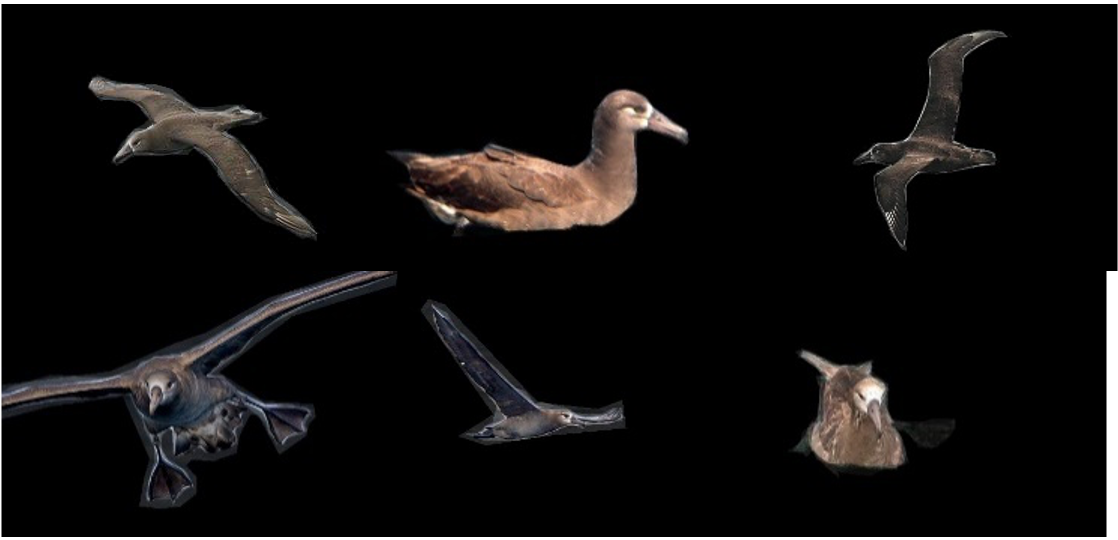}
    \caption{CUB\_Black}
    \label{fig:tbs-a}
  \end{subfigure}
  \begin{subfigure}{0.4\linewidth}
    \includegraphics[width=\linewidth]{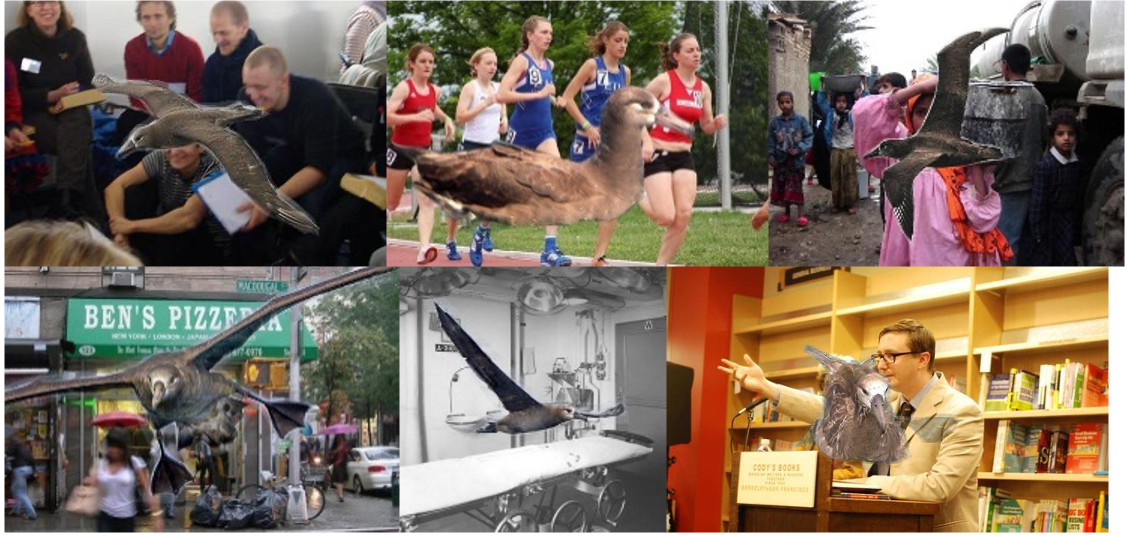}
    \caption{CUB\_Random}
    \label{fig:tbs-b}
  \end{subfigure}
  \caption{Randomly selected samples from the TravelingBirds dataset.}

  \label{fig:tbs}%
\end{figure}

Figure \ref{fig:6} presents the results of our uncertainty robustness experiments, comparing the task accuracy of our EQ-CBM with several baseline methods. In the CUB\_Black scenario, our EQ-CBM achieved the highest task accuracy, outperforming other methods. ProbCBM performed comparably but slightly lower than ours, while other methods showed reduced accuracy, indicating that our approach is more robust to background variations. Similarly, in the CUB\_Random scenario, EQ-CBM again outperformed all baseline methods. Coop was the closest competitor, but still showed slightly lower task accuracy compared to ours. Other models exhibited even lower performance, reinforcing the robustness of our model in handling diverse and unpredictable backgrounds.

Overall, the experimental results indicated that ProbCBM and ECBM are more influenced by background variations (\ie, habitat) than by object-centric concepts, whereas our EQ-CBM demonstrated superior robustness to these variations.

\begin{figure*}[t]
  \sbox0{
        \resizebox{0.47\linewidth}{!}{
          \begin{tabular}[b]{lll}
            \toprule
              & Concept \scriptsize{($\pm$CI)}               & Task \scriptsize{($\pm$CI)}\\
              \midrule
              $w/o$ EMA                 & 92.081 \scriptsize{($\pm$0.633)} &23.745 \scriptsize{($\pm$3.915)}\\
              $w/o$ Energy              & 95.454 \scriptsize{($\pm$0.167)} &78.256 \scriptsize{($\pm$0.463)}\\
              $w/o$ Var. infer.  & 96.518 \scriptsize{($\pm$0.063)} &78.681 \scriptsize{($\pm$0.433)}\\
              \rowcolor{lightgray}EQ-CBM                   & \bfseries96.580 \scriptsize{($\pm$0.043)}    & \bfseries79.310 \scriptsize{($\pm$0.272)}\\
              \bottomrule\\\\
          \end{tabular}}}%
  \begin{minipage}[b]{\wd0}
    \captionof{table}{Impact of model components on concept and task accuracy on the CUB dataset.}
    \label{tab:2}%
    \usebox0
  \end{minipage}
  \hfill
  \begin{minipage}[b]{\dimexpr \textwidth-\wd0-\columnsep}
    \stepcounter{figure}%
    \includegraphics[width=\textwidth]{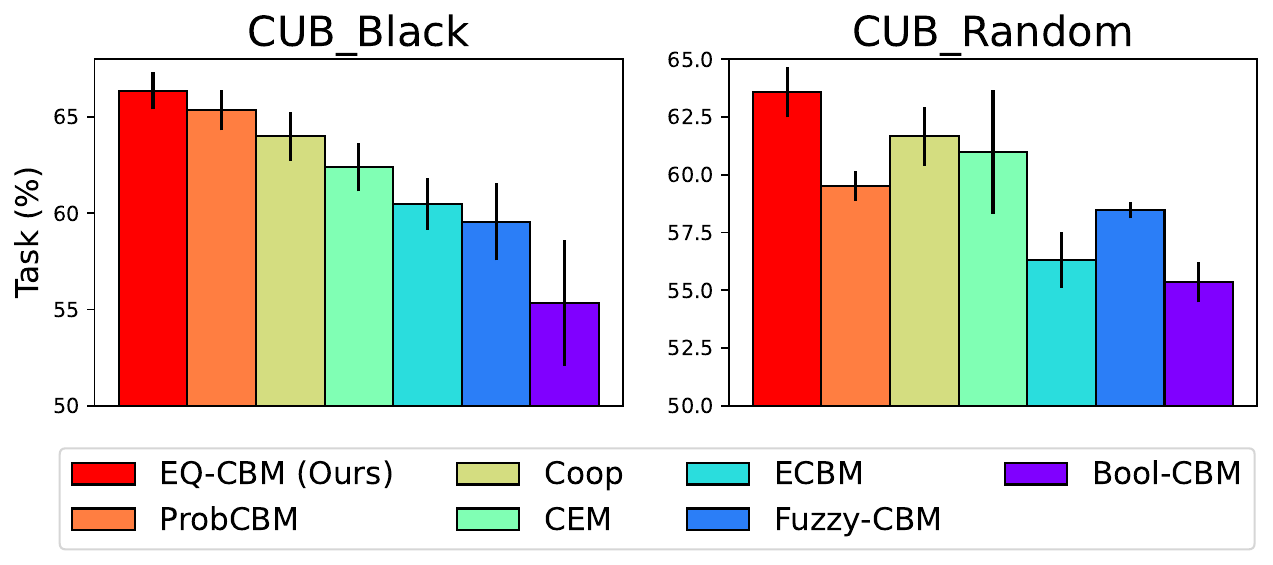}
    \addtocounter{figure}{-1}
    \vspace{-0.5cm}
    \caption{Task accuracy on the TravelingBirds dataset for CUB\_Black and CUB\_Random scenarios.}
     \label{fig:6}%
  \end{minipage}

\end{figure*}%


\subsection{Ablation Study}

  We performed an ablation study to evaluate the impact of different components of our model on both concept and task accuracy. The results are shown in Table \ref{tab:2}. Without EMA, the model’s concept accuracy dropped to 92.081\% and task accuracy significantly decreased to 23.745\%. This indicates that EMA is crucial for stabilizing and ensuring consistent convergence of qCAVs. Removing the energy-based modeling component led to a concept accuracy of 95.454\%, demonstrating that energy-based modeling is essential for capturing nuanced relationships between concepts, resulting in higher accuracy. Training without variational inference showed a slight decrease in task accuracy to 78.681\%, with concept accuracy at 96.518\%. This suggests that variational inference helps in learning diverse features, slightly enhancing performance. In contrast, the complete EQ-CBM achieved the highest performance, with a concept accuracy of 96.580\% and a task accuracy of 79.310\%, confirming the effectiveness of combining EMA, energy-based modeling, and variational inference.

\subsection{Concept Interpretation}

\begin{figure}[t]
  \centering
  \includegraphics[width=0.98\textwidth]{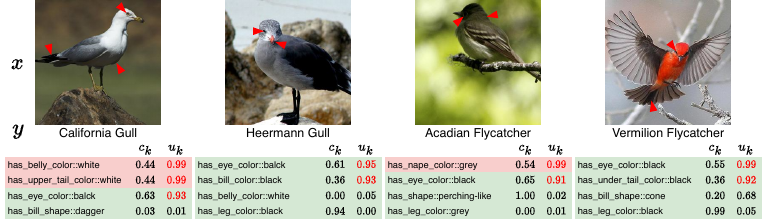}
  \caption{Interpretation of several concepts. Green boxes indicate true predictions, while red boxes indicate false predictions. $c_k$ is the concept score and $u_k$ is the uncertainty.}
  \label{fig:7}
\end{figure}

To demonstrate the interpretability of our model, we visualized several concepts and their corresponding scores and uncertainties on the CUB dataset, as shown in Fig. \ref{fig:7}. Each image depicts a bird species with its corresponding concepts, predicted concept scores $c_k$, and uncertainties $u_k$.

For the California Gull, the model incorrectly identifies the \verb|belly_color| due to shadows, resulting in high uncertainty. The \verb|upper_tail_color| also shows high uncertainty because it is occluded. The \verb|eye_color| is correctly identified but with high uncertainty due to its multicolored nature. For the Heermann Gull, the \verb|eye_color| and \verb|bill_color| are correctly identified but have high uncertainty because of the small size of the eyes and the multicolored bill. In the Acadian Flycatcher image, the \verb|nape_color| prediction has some uncertainty due to a shadow that makes the nape appear gray. Although the predicted concept is incorrect, the uncertainty score indicates that the model is unsure about this prediction. Similarly, the \verb|eye_color| prediction also shows high uncertainty due to the shadow effect. For the Vermilion Flycatcher, the model predicts the \verb|under_tail_color| with some uncertainty. While the \verb|under_tail| is not entirely black, the presence of some black in the actual tail leads to relatively high uncertainty. The \verb|eye_color| prediction also shows high uncertainty because the bird is facing forward, making it difficult for the model to judge accurately.

These visualizations help in understanding the model’s decision-making process, highlighting where the model is less confident. This interpretability is crucial for validating the model’s predictions and identifying areas for potential improvement.


\section{Conclusion}\label{sec:5}

In this paper, we introduced EQ-CBM, a novel framework designed to enhance CBMs through probabilistic concept encoding using EBMs with qCAVs. Our approach addressed the limitations of deterministic concept encoding in existing CBMs by enabling robust probabilistic inference, thereby improving both concept and task accuracy. By integrating EBMs, our framework effectively captured the underlying dependencies and uncertainties within the data, leading to more accurate and interpretable models. qCAVs ensured the selection of homogeneous vectors during concept encoding, enhancing both task performance and human intervention capabilities. Experiments on datasets such as CUB-200-2011, AwA2, CelebA, and TravelingBirds demonstrated the superiority of EQ-CBM in achieving a better balance between interpretability and accuracy compared to existing CBMs. The TravelingBirds dataset further showcased the robustness of our model under challenging conditions. Our ablation studies underscored the importance of each component within our framework.

Future work will focus on improving the scalability of our approach to efficiently handle larger datasets, providing a robust and flexible framework for enhancing the decision-making processes of DNNs.

\section*{Acknowledgement}

This research was supported by Electronics and Telecommunications Research Institute (ETRI) grant funded by the Korean government. [24ZD1120, Development of ICT Convergence Technology for Daegu-GyeongBuk Regional Industry].


%
%
\bibliographystyle{splncs04}
\bibliography{main}

\clearpage
\section*{Appendix}
\renewcommand{\thetable}{A\arabic{table}}
\renewcommand{\thefigure}{A\arabic{figure}}
\renewcommand\thesection{\Alph{section}}
\setcounter{section}{0}
\setcounter{figure}{0}
\setcounter{table}{0}

\section{Model Complexity}

In this section, we present a detailed comparison of the model complexity for various concept bottleneck models (CBMs), including our proposed EQ-CBM. Table \ref{tab:complexity} shows the number of parameters, FLOPs, and inference latency for each method. Additionally, it highlights the concept accuracy and task accuracy. These metrics emphasize the computational efficiency and performance of our approach relative to existing methods.

\setlength{\tabcolsep}{6pt}
\begin{table}
\centering
\caption{Model complexity comparison across different CBM methods on the CUB dataset.}
\label{tab:complexity}
\begin{tabular}{lccccc}
\toprule
Methods & \#Params& FLOPs  & Latency  & Concept & Task  \\
        & (M)  $\downarrow$         & (G) $\downarrow$        & (ms) $\downarrow$       & (\%) $\uparrow$             & (\%) $\uparrow$       \\
\midrule
Fuzzy-CBM \cite{CBM} & 21.36 & 6.85  & 4.86 & 95.882 & 74.228\\
Coop-CBM \cite{COOP}  & 21.47 & 6.84  & 5.86 & 89.892 & 79.154\\
Prob-CBM \cite{prob} & 22.65 & 7.38  & 49.38 & 95.596 & 76.265\\
ECBM \cite{ECBM}  & 23.34 & 6.84  & 23.17 & 96.536 & 77.148\\
CEM \cite{CEM} & 23.88 & 6.85  & 8.74 & 96.160 & 79.029\\
\rowcolor{lightgray}EQ-CBM (Ours)  & 23.75 & 6.85 & 4.86 & 96.580 & 79.310\\
\bottomrule
\end{tabular}
\end{table}

The results highlighted that EQ-CBM struck a balance between model complexity and performance. Specifically, EQ-CBM had 23.75M parameters, comparable to CEM but higher than Fuzzy-CBM and Coop-CBM. In terms of computational requirements, EQ-CBM had similar GFLOPs to most other models, indicating efficient computational performance. Despite the additional complexity, EQ-CBM matched the inference latency of the original CBM at 4.86 ms, ensuring that the enhancements did not compromise speed. Moreover, EQ-CBM achieved superior concept and task accuracy compared to other models. It led in both metrics, with a concept accuracy of 96.580\% and a task accuracy of 79.310\%. This demonstrated the effectiveness of our method in improving both interpretability and task accuracy.

In contrast, while other models such as Prob-CBM and ECBM showed substantial computational requirements and latency, they fell short in terms of task accuracy compared to EQ-CBM. For instance, Prob-CBM had a significantly higher latency of 49.38 ms, highlighting inefficiencies in practical scenarios despite its competitive concept accuracy. These findings underscored that EQ-CBM provided an optimal balance between computational efficiency and high performance.

\section{Nearest Neighbor Concept Analysis}

\begin{figure}[h]
\includegraphics[width=\linewidth]{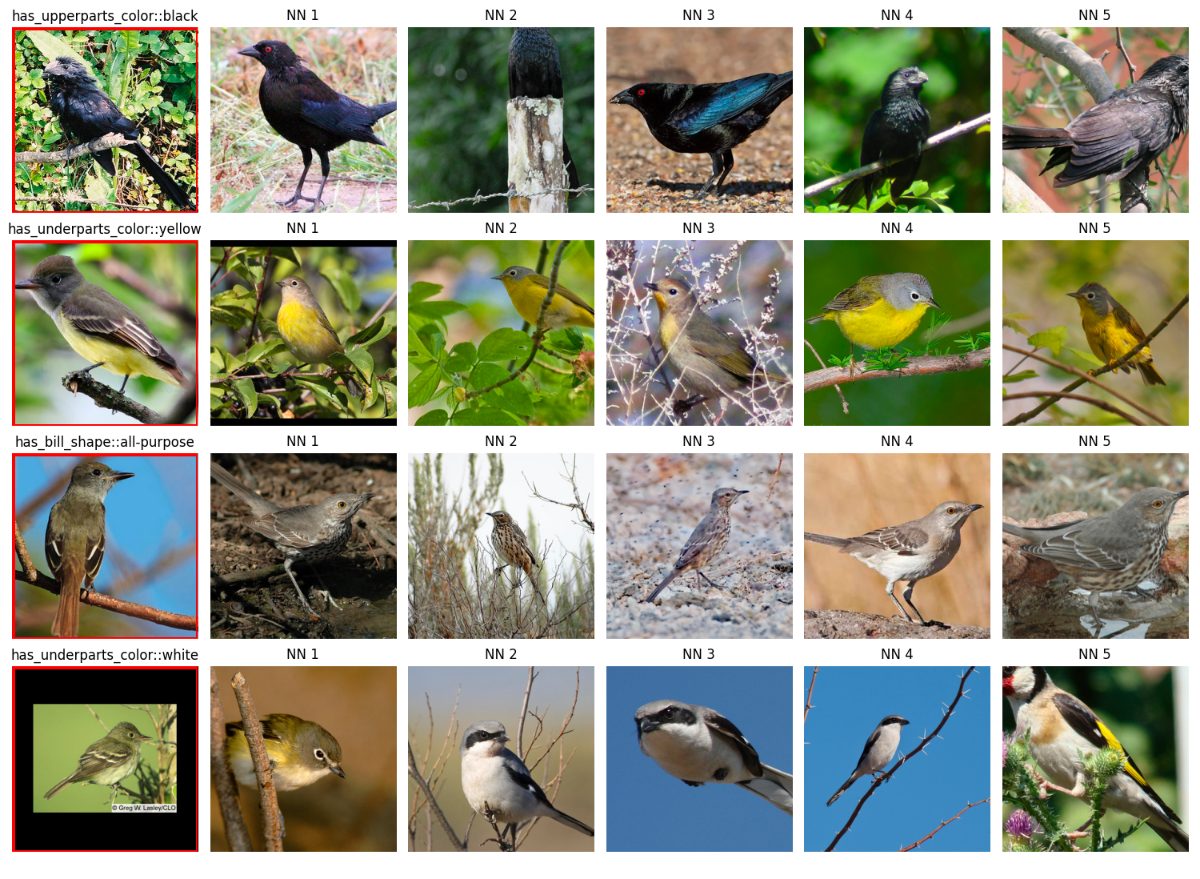}
\caption{Nearest neighbor analysis for selected bird concepts. Each row shows a query image (left, with red border) and its top-5 nearest neighbors for the specified concept.}
\label{fig:nn_concept_analysis}
\end{figure}

To further validate the interpretability and robustness of our proposed EQ-CBM model, we conducted a nearest neighbor (NN) concept analysis. This experiment visualized the top-5 NNs for selected concept embeddings, allowing us to assess how well the model captured and represented these concepts across different images. Figure \ref{fig:nn_concept_analysis} illustrates the NN results for four selected concepts. For each concept, the query image (highlighted with a red border) is shown on the left, followed by its top-5 NN images identified by the model.

The NN analysis visually confirmed EQ-CBM’s capability to effectively learn and represent distinct bird concepts. The consistent retrieval of relevant images across different concepts underscored the robustness and interpretability of our approach. These results suggested that EQ-CBM could reliably capture complex visual attributes, enhancing both the transparency and performance of CBMs.

\section{Energy Distribution Analysis}

\begin{figure}[h]
\centering
\includegraphics[width=\linewidth]{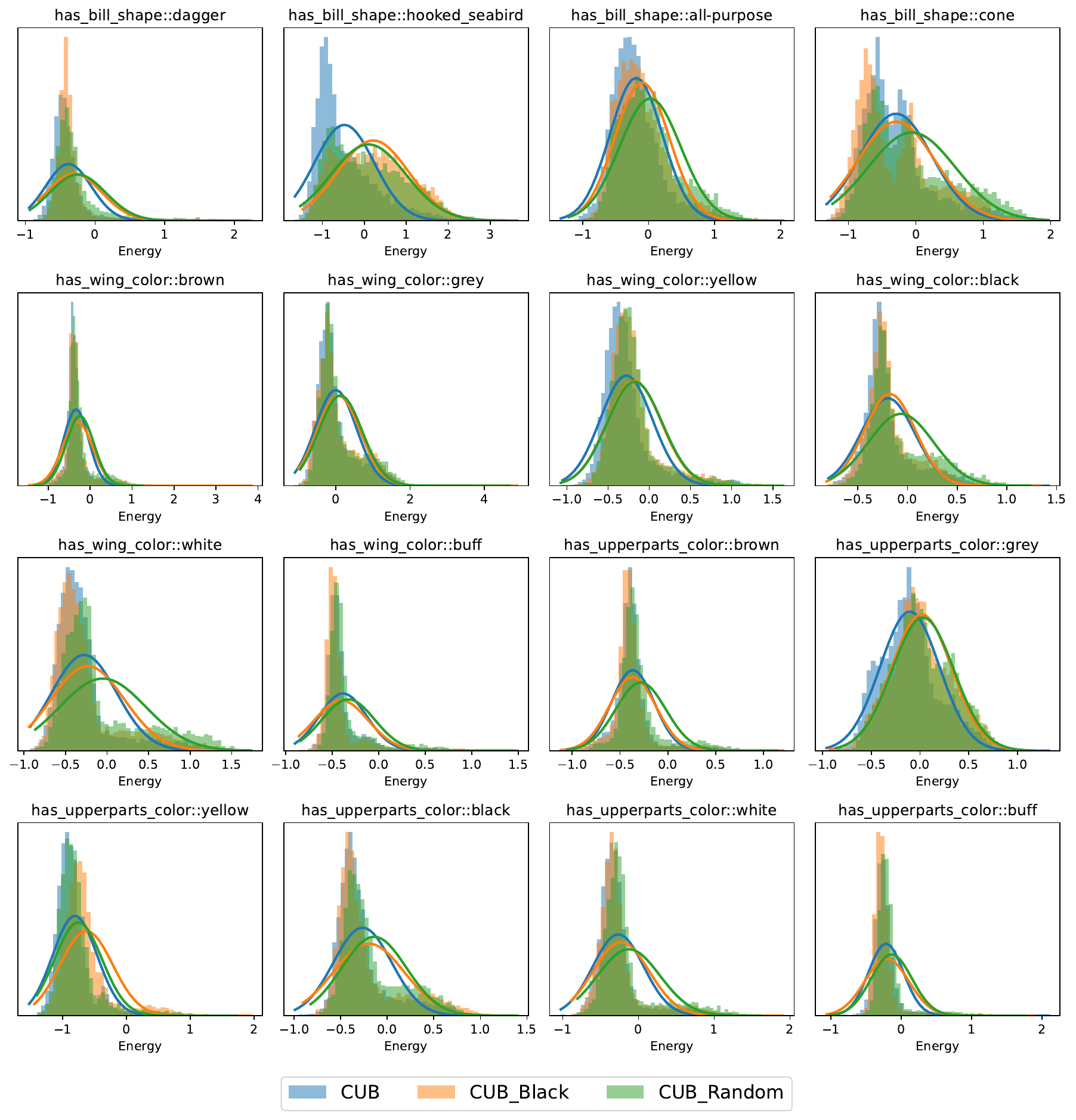}
\caption{Each subplot shows the energy distribution for a specific concept.}
\label{fig:energy_dist}
\end{figure}

We present an analysis of the energy distributions generated by EBMs in our probabilistic concept encoders for various bird concepts. Figure \ref{fig:energy_dist} illustrates the energy ($\bar{e}_k$) distributions for selected bird concepts across three different datasets: CUB, CUB\_Black, and CUB\_Random. The CUB dataset \cite{welinder2010caltech} is the original data used for training, while CUB\_Black and CUB\_Random are subcategories of the TravelingBirds dataset \cite{CBM} used solely for inference, where the backgrounds were replaced with black and random real-world scenes \cite{Places}, respectively.

The energy distributions revealed that our EQ-CBM model maintained uniform probability densities across the original and modified datasets. This uniformity indicated that the model’s performance was robust to background variations, as the energy values for the same concepts remained consistent regardless of changes in the background. For instance, the concept \texttt{has\_bill\_shape::dagger} showed a similar energy distribution across CUB, CUB\_Black, and CUB\_Random datasets, highlighting the model’s ability to consistently capture this concept despite varying backgrounds. Similarly, other concepts such as \texttt{wing\_color} exhibited comparable energy distributions across different datasets, further validating the robustness of our model.

The uniformity of energy distributions across different datasets emphasized the effectiveness of the proposed probabilistic concept encoders with EBMs and qCAVs. The ability to maintain consistent probability densities across the original and modified datasets ensured that our EQ-CBM could accurately distinguish between different concepts, leading to high interpretability and task accuracy. This robustness is crucial for real-world applications where models often encounter diverse and unpredictable scenarios.

\end{document}